\documentclass{article}
\usepackage{spconf,amsmath,graphicx}
\usepackage{amsmath,amsfonts}
\usepackage{algorithmic}
\usepackage{algorithm}
\usepackage{array}
\usepackage{subfigure}
\usepackage{graphicx}
\usepackage{multicol}
\usepackage{multirow}
\usepackage{textcomp}
\usepackage{stfloats}
\usepackage{url}
\usepackage{verbatim}
\usepackage{graphicx}
\usepackage{cite}
\usepackage{booktabs}
\usepackage{tabularx}
\usepackage{amsfonts,amssymb}
\usepackage{multirow}
\usepackage{amsmath,amsfonts}
\usepackage{algorithmic}
\usepackage{algorithm}
\usepackage{array}
\usepackage{bbding}
\usepackage{wrapfig}
\usepackage[table]{xcolor}
\newcommand{\gr}{\rowcolor[gray]{.95}}

\title{MSP-Former: Multi-Scale Projection Transformer for Single Image Desnowing}
\name{Sixiang Chen$^{1\dag}$, Tian Ye$^{1\dag}$, Yun Liu$^{2\dag}$, Taodong Liao$^{1}$, Jingxia Jiang$^{1}$, Erkang Chen$^{1,3}$, Peng Chen$^{1,3*}$.
\thanks{$^{1}$Sixiang Chen, Tian Ye, Taodong Liao, JingXia Jiang, Erkang Chen and Peng Chen are with the School of Ocean Information Engineering, Jimei University, Xiamen, China. E-mail: \{\tt\small 201921114013, 201921114031, 201921114028, 202021114006, ekchen, chenpeng\}@jmu.edu.cn}%
\thanks{$^{2}$Yun Liu is with the College of Artificial Intelligence, Southwest University, Chongqing, China. 
        {\tt\small yunliu@swu.edu.cn}}%
\thanks{$^{3}$Erkang Chen and Peng Chen are with the Fujian Provincial Key Laboratory of Oceanic Information Perception and Intelligent Processing.}%
\thanks{This research was supported by  Natural Science Foundation of Fujian Province, China (2021J01867), Natural Science Foundation of Chongqing, China (cstc2020jcyj-msxmX0324), Xiamen Municipal Bureau of Ocean Development (22CZB013HJ04).}
\thanks{$^{*}$: Corresponding author. $\dag$: Equal Contribution}
}

\address{}
\vspace{-1.5cm}
\begin{document}

\small
%
\maketitle
\begin{abstract}
Snow removal causes challenges due to its characteristic of complex degradations. To this end, targeted treatment of multi-scale snow degradations is critical for the network to learn effective snow removal. In order to handle the diverse scenes, we propose a multi-scale projection transformer (MSP-Former), which understands and covers a variety of snow degradation features in a multi-path manner, and integrates comprehensive scene context information for clean reconstruction via self-attention operation. For the local details of various snow degradations, the local capture module is introduced in parallel to assist in the rebuilding of a clean image. Such design achieves the SOTA performance on three desnowing benchmark datasets while costing the low parameters and computational complexity, providing a guarantee of practicality.
\end{abstract}

\section{Introduction}
With the development of deep learning, low-level vision tasks have made great progress~\cite{pmnet,Ye_2022_CVPR,jin2022unsupervised,zamir2021restormer,ye2022mutual}.
Snow removal from a single image is a complicated vision task because multiple ill-conditioned degradations (e.g. veiling effect, background occlusion, snow particles etc.) are superimposed on a snowy image. The imaging model of the snow scene can be modeled as:
\begin{equation}\label{snow}
    I(x) = K(x)T(x) + A(x)(1-T(x)),
\end{equation}
where $I(x)$ represents the snowy image, $K(x)=J(x)(1-Z(x)R(x)+C(x)Z(x)R(x)$. $K(x)$ is the veiling-free snowy image, $T(x)$ and $A(x)$ are the transmission map and atmospheric light. $C(x)$ and $Z(x)$ denote the chromatic aberration map for snow images and the snow mask, respectively. $J(x)$ is the snow-free image.

Earlier single image desnowing methods were based on the priors~\cite{pei2014removing,wang2017hierarchical,rajderkar2013removing,zheng2013single,Ye_2022_ACCV,chen2022snowformer}. For instance, Pei~\emph{et al}.~\cite{pei2014removing} utilized the image features prior of visibility and saturation to remove snowflakes. Recently, with the development of CNNs, several desnowing networks~\cite{liu2018desnownet,allinone,chen2020jstasr,hdcwnet,zhang2021deep} had been proposed to restore the clean image. JSTASR~\cite{chen2020jstasr} employed a divide and conquer strategy to deal with the degradation of different sizes and haze veiling, respectively. In HDCW-Net~\cite{hdcwnet}, the dual-tree complex wavelet transform was applied to decompose the snow scene image. The hierarchical high and low frequency reconstruction network is used to remove the snow degradations and reconstruct the clean scene. DDMSNet~\cite{zhang2021deep} considered the semantic and depth information to precisely promote the network to tackle snow degradations. 

\textbf{Motivation}. Despite that, two crucial points for effective desnowing are ignored: (i). \textit{Targeted processing of multi-scale snow degradations and the relationship between them are the basis for handling complex snow scenes.} (ii). \textit{Global context information is indispensable for the perfect reconstruction of clean scenes, which is overlooked in previous methods.} From Eq.\ref{snow}, the degraded snow scene usually contains the snowflake, snow streak, and veiling effect. Various degradations have different scale characteristics, making it easy to ignore some degradation information by fixing a single desnowing paradigm. In addition, the global snow scenes interaction cannot be acquired sufficiently via simple semantic and depth priors. 
\begin{figure}[!t]

\centering
\includegraphics[width=0.47\textwidth]{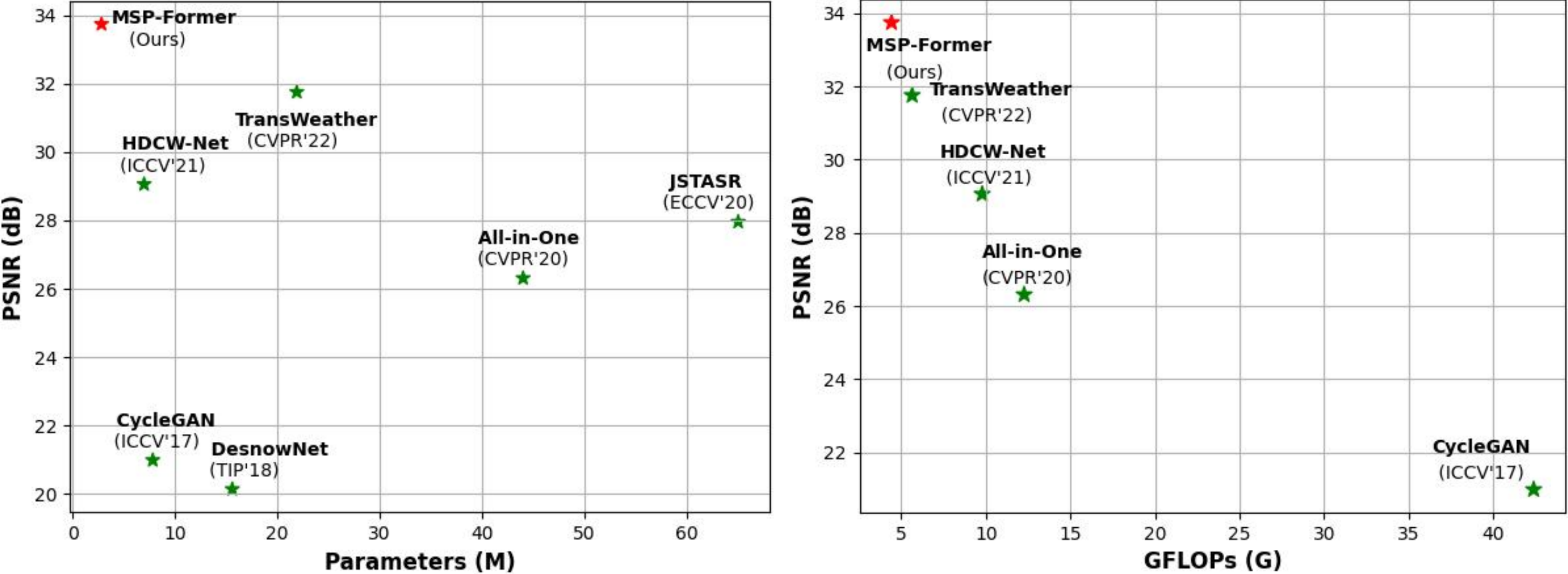} 
\caption{\small 
\textbf{Left}. Trade-off between performance vs: number of parameters on CSD~\cite{hdcwnet} testing dataset. \textbf{Right}. Trade-off between performance vs: computational cost calculated on $256 \times 256$ resolution.}\label{gflop}

\end{figure}


To settle above problems, we propose the MSP-Former, which particularly targets diverse snow degradations and global context information via a multi-scale self-attention paradigm. And the local snow degradation is also fully considered.
Specifically, we construct a channel-dimensionally parallel network architecture, including a multi-scale projection (MSP) and a local capture module (LCB). MSP adopts the average pooling operation with different kernels and strides to aggregate diverse snow scenes and reduces the feature size. Furthermore, we separately project them into Key ($\mathbf{K}$) and Value ($\mathbf{V}$) for the multi-scale self-attention mechanism with global context information.
Such a compact design leads our network to achieve SOTA performance at a much lower computational and parameter amount than previous ones, as shown in Fig.\ref{gflop}.

More concretely, our contributions are summarized as follows:
\begin{itemize}
    \item We propose an effective and lightweight single image snow removal architecture, MSP-Former, which is set as two parallel branches along the channel dimension. One target focuses on diverse snow degradations to support for the reconstruction of clean scenes and another branch aims to refine local details for snow scenes. 
    \item Multi-scale Projection module primarily aggregates various snow degradations, which interacts with the global snow context information via multi-head self-attention to enhance the representation of reconstructing a clean image.
    \item We conduct extensive experiments to verify the superior performance of the proposed method. The experiments demonstrate that our network outperforms the previous state-of-the-art methods on multiple snow removal datasets, while the computational overhead and parameter amount are also significantly superior to previous methods.
\end{itemize}

\vspace{-0.6cm}

\section{Method}
\begin{figure}[!t]
\setlength{\abovecaptionskip}{0cm} 
\setlength{\belowcaptionskip}{-0.7cm}
\centering
\includegraphics[width=85mm]{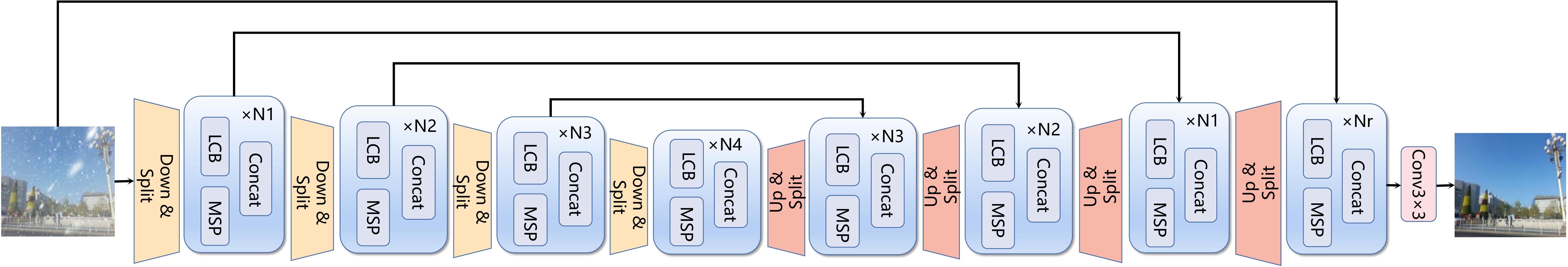} 
\caption{\small 
Our MSP-Former consists of an encoder-decoder, and there is a skip connection between the corresponding stages inspired by UNet~\cite{unet}. Ultimately, we use a $3\times3$ convolution to transform the output image.}
\label{fig:overview}
\end{figure}

\vspace{-0.2cm}
An overview of the MSP-Former architecture is shown in Fig.\ref{fig:overview}, which is a multi-scale encoder and decoder architecture followed by UNet~\cite{unet}. We employ $3\times3$ convolutional downsampling for each stage to reduce image size and increase dimension number. The features are split into two parts along the channel dimension and feed into our well-designed MSP and LCB modules for global clean scene reconstruction and local feature capture, respectively. We perform a channel shuffle operation before each channel concatenate, and ablation experiments prove that the channel shuffle operation can optimize our model. Afterward, we integrate the separately processed information along the channel dimension to improve the representative ability of our model on the image snow removal task. In the final stage, we add a refinement module based on the above design.

\vspace{-0.4cm}
\subsection{MSP Module}
\vspace{-0.2cm}
In snowy images, the ill-conditioned snow degradations of different scales overlaid on clean images make snow removal a challenging problem. To address this issue, we elaborate on a multi-scale projection (MSP) module to handle various degradations, which combines the global snow context information via self-attention.
\vspace{-0.3cm}
\subsubsection{MSP Self-Atention}
\begin{figure*}[!t]
\centering
\includegraphics[width=112mm]{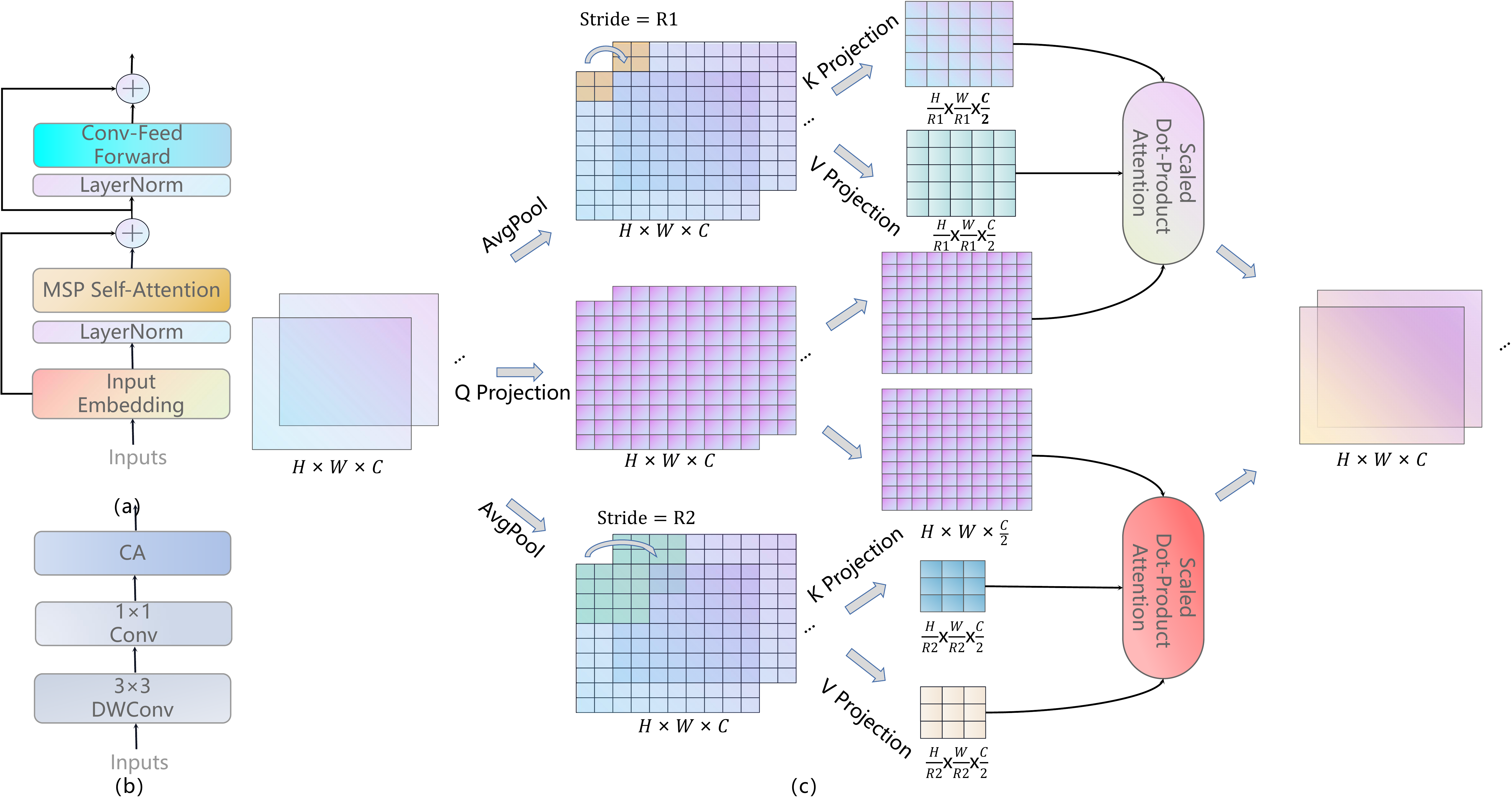} 
\caption{\small (a) is the multi-scale projection(MSP) module, (b) is the local capture block(LCB) module, and (c) is the multi-scale projection self-attention mechanism(MSP Self-Attention) in the MSP module.}
\label{fig:msp-former}
\end{figure*}

\vspace{-0.2cm}
For the vision transformer, as shown in Fig.\ref{fig:msp-former}, given an input feature $X^{H\times W\times C }$, it is usually reshaped into a 3D sequence $X_{s}^{N\times C}$ and a linear layer is employed to project $X$ into Query($\mathbf{Q}$), Key($\mathbf{K}$), Value($\mathbf{V}$), can be expressed in the following form:
\begin{equation}
    \mathbf{Projection:}\; \mathbf{Q} = W_{q}X_{s}, \mathbf{K} = W_{k}X_{s}, \mathbf{V} = W_{v}X_{s},
\end{equation}
where $W_{q}$, $W_{k}$, $W_{v}$ are linear projection parameters. Nevertheless, due to fixed patch embedding, separate projections result in a fixed receptive field for each sequence in the global modeling after linear layer transformation. This is very disadvantageous for recovering variable degradation information in complex snow images.

To solve this problem, in MSP self-attention, we first use a multi-scale avgpool to aggregate different information, fully capturing diverse scale snow information and providing multi-head self-attention with a richer snow degradation representation capability. At the same time, based on the above operations, we reduce the image size required for K and V projection to reduce the amount of self-attention computation, and we can express the aggregated features as:
\begin{equation}
    X_{1}^{\frac{H}{R_1}\times \frac{W}{R_1}\times C } = \operatorname{Avgpool}^{S=R_1}_{K1}(X^{H\times W\times C }),
\end{equation}
\begin{equation}
    X_{2}^{\frac{H}{R_2}\times \frac{W}{R_2}\times C } = \operatorname{Avgpool}^{S=R_2}_{K2}(X^{H\times W\times C }),
\end{equation}
among them, $X_1$ and $X_2$ are different feature maps, and K and S represent pooled kernel and stride. Then we use a linear layer to re-project the features that aggregate different local information to Key($\mathbf{K}$) and Value($\mathbf{V}$), respectively, for Scaled Dot-product Attention with the original Query($\mathbf{Q}$) to improve the multi-scale representation ability of self-attention. Such a paradigm incorporates global context information and multi-scale degraded snow scenes into the design category, which carefully considers the critical motivation for snow removal.
Inspired by multi-head self-attention~\cite{dosovitskiy2020image}, we perform multi-head self-attention (MSA) separately for Key($\mathbf{K}$) and Value($\mathbf{V}$) with different degradations, and finally concat them together. The formula of the multi-scale projection can be expressed as follows:
\begin{equation}
     K_{\left\{1,2\right\}} = W_{k\left\{1,2\right\}}X_{\left\{1,2\right\}}, V_{\left\{1,2\right\}} = W_{v\left\{1,2\right\}}X_{\left\{1,2\right\}},
\end{equation}
\begin{equation}
    \mathbf{MSA}(X) = {\operatorname{Concat}}(\operatorname{Softmax}\left(\frac{\mathbf{Q_{h}} \mathbf{K_{h\left\{1,2\right\}}}^{\top}}{\sqrt{D_{h}}}\right) \mathbf{V_{h\left\{1,2\right\}}}),
\end{equation}
where h represents multi-head processing, and $D_{h}$ is the dimension of each head. Considering the fine information and low-level details required in the process of self-attention, we add a depth-wise convolution before the point product of Value($\mathbf{V}$) to improve the local retention ability like \cite{ren2022shunted} does.

\vspace{-0.4cm}
\subsubsection{MSP Transformer Block}
Like the vanilla Transformer, the MSP transformer block consists of two critical constructions, MSP Self-Attention and ConvFFN. It adds Layernorm before each part while adding residual connections to the MSP Self-Attention and ConvFFN parts to stabilize the training network. The following formula can describe the entire block:
\begin{equation}
\begin{aligned}
&{X^{'}}={X}+\operatorname{\mathbf{MSP-Self-Attention}}(\operatorname{\mathbf{LN}}({X})), \\
&{Y}={X'}+\operatorname{\mathbf{ConvFFN}}(\operatorname{\mathbf{LN}}({X'})),
\end{aligned}
\end{equation}
We continue to use ConvFFN~\cite{PVT}, which has been demonstrated to possess more potential to replace the traditional MLP feedforward network to alleviate its shortcomings in local modeling, similar to the PVT\cite{PVT} approach. 
\vspace{-0.3cm}
\subsection{LCB Module}
In the single snow removal task, both the global degradation and the local degradation consisting of snow patches are included, so the local context information determines the detailed features of the restored image. Unlike the multi-scale self-attention module introduced above, we develop a local capture module parallel to the MSP module. It is worth mentioning that it differs from the complex calculation required by the transformer module. It only requires minimal computation and parameters, which plays a crucial role in the model's performance-parameters trade-off in the snow removal task.

As illustrated in Fig.2(b), for the input $X^{H\times W\times C }$, we utilize $3\times 3$ depth-wise convolution to excavate local snow features and exploit a $1\times 1$ convolution for inter-channel interactions. In addition, the transformer is essentially self-modeling at the spatial level between patches. It lacks the modeling between channels, so we add channel attention~\cite{hu2018squeeze} to the local capture module to increase the overall representation ability of the model. We can compute the whole LCB module as follows:
\begin{equation}
    LCB(X^{H\times W\times C }) = \operatorname{CA}(\operatorname{Conv}_{K=1}(\operatorname{DWConv}_{K=3}(X))),
\end{equation}
where $K$ is the convolution kernel and CA is the channel attention. 
\vspace{-0.4cm}
\subsection{Loss Function}
We introduce the Charbonnier Loss~\cite{charbonnier1994two} as our reconstruction loss:
\begin{equation}
    \mathcal{L}_{rec} = \mathcal{L}_\mathcal{CR}(\mathcal{M}(X)-\mathcal{J}_{gt}),
\end{equation}
where the $\mathcal{M}$ is proposed network, $X$ and $\mathcal{J}_{gt}$ stand for input and ground-truth. $\mathcal{L_{CR}}$ denotes the Charbonnier loss, which can be express as:
\begin{equation}
    \mathcal{L_{CR}} =\frac{1}{N} \sum_{i=1}^{N} \sqrt{\left\|X^{i}-Y^{i}\right\|^{2}+\epsilon^{2}},
\end{equation}
where constant $\epsilon$ emiprically set to 1e$^{-3}$ for all experiments.
\vspace{-0.3cm}
\section{Experiments}

\vspace{-0.2cm}
\noindent\textbf{Implementation Details}.
In the training details of the proposed MSP-Former, we use AdamW optimizer with the first and second momentum terms of (0.9, 0.999), the weight decay is 5e$^{-4}$ to train our framework. We set the initial learning rate to 0.0007 and employ linear decay after 250 epochs. For data augmentation, we adopt horizontal flip and random rotation by 90, 180, 270 degrees. We randomly crop a 256$\times$ 256 patch as input to train 600 epochs.
\begin{table}[!t]
\centering
\setlength{\abovecaptionskip}{0.5cm}
\caption{\small Desnowing results on the CSD~\cite{hdcwnet}, SRRS~\cite{chen2020jstasr} and Snow 100K~\cite{liu2018desnownet} datasets (PSNR(dB)/SSIM). Bold and underline indicate the best and second metrics. }\label{snowresults}
\resizebox{8cm}{!}{
\renewcommand\arraystretch{1.1}
\begin{tabular}{l|cc|cc|cc|c|c}
\toprule[1.2pt]
\multirow{2}*{Method}& \multicolumn{2}{c|}{ CSD (2000)~\cite{hdcwnet} } & \multicolumn{2}{c|}{ SRRS (2000)~\cite{chen2020jstasr} } & \multicolumn{2}{c|}{ Snow 100K (2000)~\cite{liu2018desnownet} } & \multirow{2}{*}{\#Param} & \multirow{2}{*}{\#GMacs}\\\cline{2-7}
& PSNR $\uparrow$ & SSIM $\uparrow$ & PSNR $\uparrow$ & SSIM $\uparrow$ & PSNR $\uparrow$ & SSIM $\uparrow$\\
\hline
(TIP'18)DesnowNet~\cite{liu2018desnownet} & 20.13 & 0.81 &20.38 &0.84& 30.50 & 0.94 & {15.6M}& 1.7KG\\
(CVPR'18)CycleGAN~\cite{engin2018cycle}& 20.98 & 0.80 &20.21 &0.74& 26.81 & 0.89 & 7.84M& {42.38G}\\
(CVPR'20)All in One~\cite{allinone} &26.31 &{0.87} &24.98 &0.88& 26.07&0.88 & 44M& { 12.26G}\\
(ECCV'20)JSTASR~\cite{chen2020jstasr} &27.96 &0.88 & 25.82 & 0.89 & 23.12 & 0.86 & {65M}& $\mathbf{-}$\\
(ICCV'21)HDCW-Net~\cite{hdcwnet} & {29.06} &{0.91} &{27.78} &{0.92} & {31.54} &\underline{0.95} & 6.99M& {9.78G}\\
(CVPR'22)TransWeather~\cite{valanarasu2022transweather} &$\underline{31.76}$ &$\underline{0.93}$ &\underline{28.29} &\underline{0.92} &\underline{31.82} &{0.93} & 21.9M & {5.64G}\\
\hline\hline 
\gr  MSP-Former &$\mathbf{33.75}$&$\mathbf{0.96}$ & $\mathbf{\mathbf{30.76}}$ &$\mathbf{0.95}$& $\mathbf{33.43}$ & $\mathbf{0.96}$ & 2.83M& {4.42G}\\
\bottomrule[1.2pt]
\end{tabular}}
\end{table}
For a concise description, we illustrate our MSP-Former framework design. At each stage of the network, we set the channel dimensions to $\left\{32,64,128,256\right\}$ respectively, and we gradually increase the number of MSP and LCB modules to $N_{1}$, $N_ {2}$, $N_{3}$, $N_{4}$, which are set to 2, 3, 4, and 6 in our network. We specify the strides $R_{1}$ and $R_{2}$ of each layer's multi-scale average pooling as $R_{1}$ = $\left\{16,8,4,2\right \}$ and $R_{2}$ = $\left\{8,4,2,1\right\}$, the pooling kernel is also set as $K_{1}$ = $\left\{16,8,4,2\right \}$, $K_{2}$ = $\left\{8,4,2,1\right\}$. At the shallowest layer of the decoder, we also add an MSP and an LCB module to refine the original resolution image to improve the final output, in which $N_{r}$, $K_{1}, R_{1}$ and $K_{2}, R_{2}$ are set to 1, 16, 16, 8, 8.


\noindent\textbf{Quantitative Analysis}. In this section, we compare the performance of our method with other existing snow removal methods: \cite{liu2018desnownet}, \cite{engin2018cycle}, \cite{allinone}, \cite{chen2020jstasr}, \cite{hdcwnet} and \cite{valanarasu2022transweather} on three benchmarks~\cite{liu2018desnownet,chen2020jstasr,hdcwnet}. To ensure fairness, for the methods without snow datasets training, we re-train to meet the latest evaluation criteria of the previous snow removal~\cite{hdcwnet} and take the best training results for comparison. Table \ref{snowresults} reports the results on different snow datasets, which shows that our model outperforms all SOTA desnowing methods. On Snow100K~\cite{liu2018desnownet}, SRRS~\cite{chen2020jstasr} and CSD~\cite{hdcwnet} testing sets, MSP-Former achieves $1.99$ dB, $2.47$ dB and $1.61$ dB improvement on PSNR metric, and is also significantly ahead of the most advanced methods on SSIM.

\begin{figure*}[t!]
    \vspace{-1mm}
    \begin{center}
        \begin{tabular}{ccccccc}
\includegraphics[width = 0.1\linewidth]{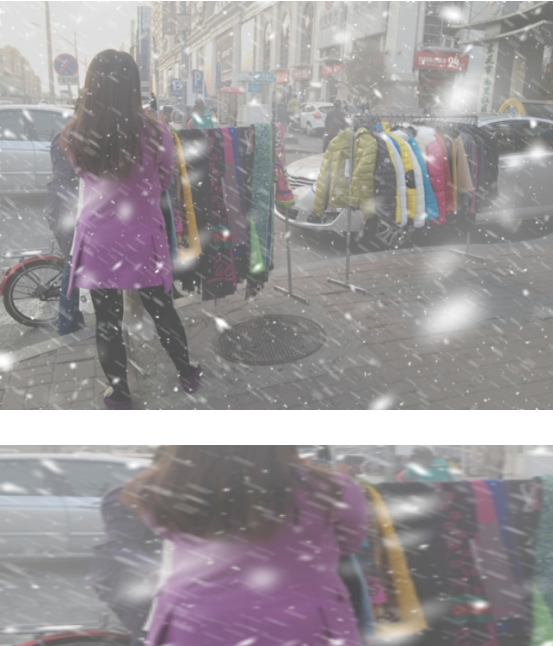} &\hspace{-4mm}
\includegraphics[width = 0.1\linewidth]{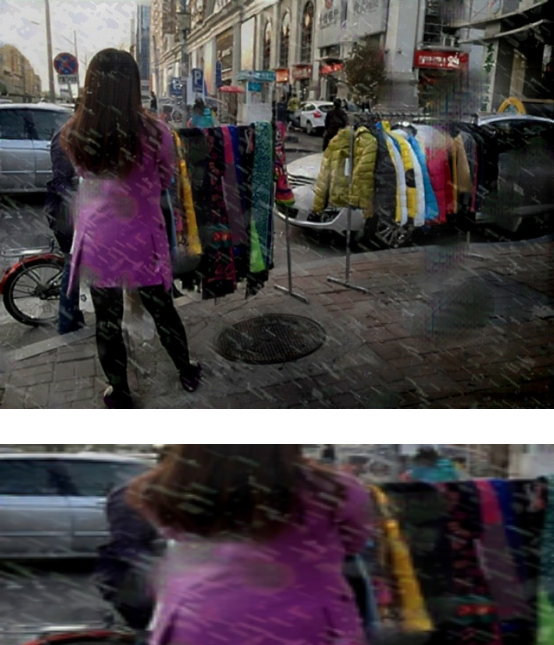} &\hspace{-4mm}
\includegraphics[width = 0.1\linewidth]{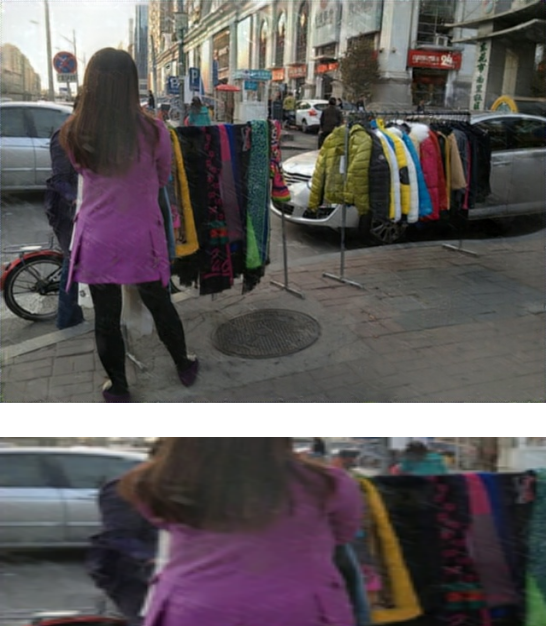} &\hspace{-4mm}
\includegraphics[width = 0.1\linewidth]{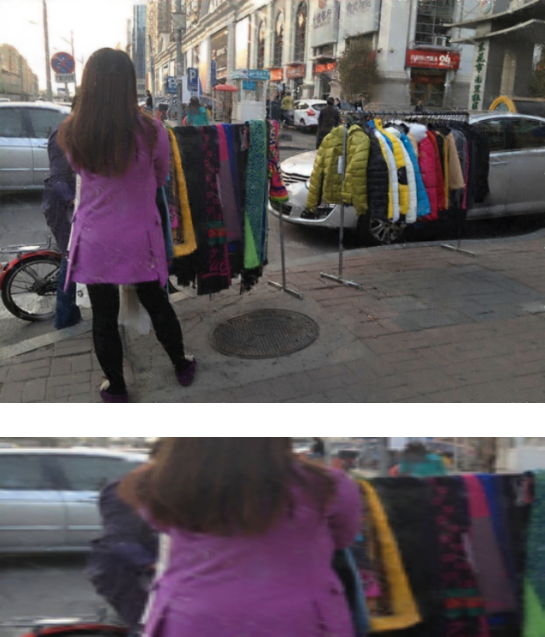} &\hspace{-4mm}
\includegraphics[width = 0.1\linewidth]{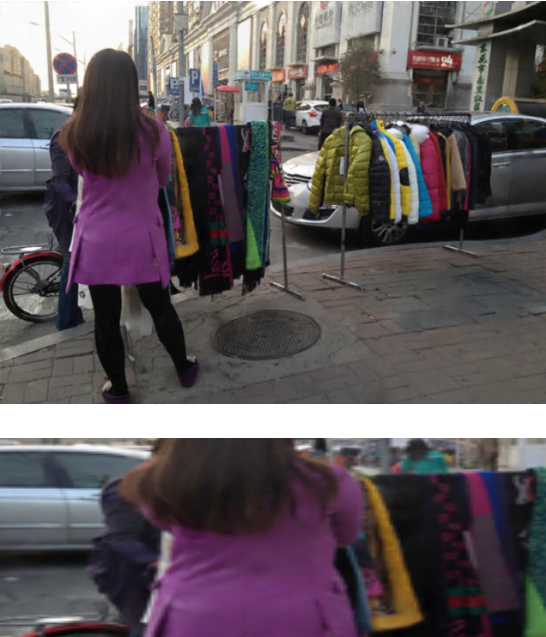} &\hspace{-4mm}
\includegraphics[width = 0.1\linewidth]{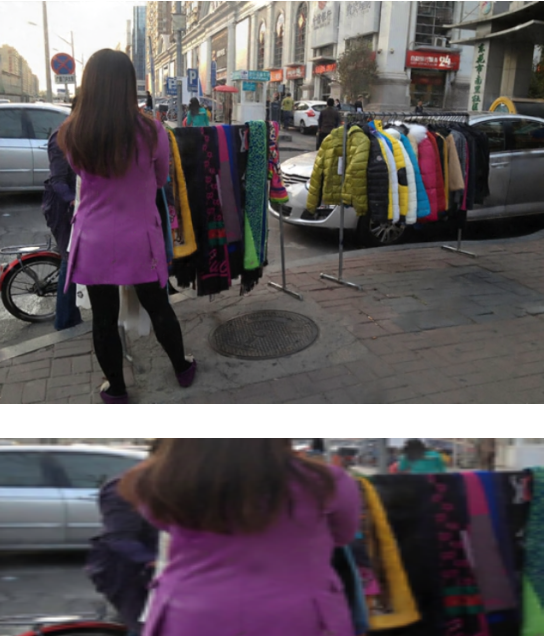} & \hspace{-4mm}
\\
\includegraphics[width = 0.1\linewidth]{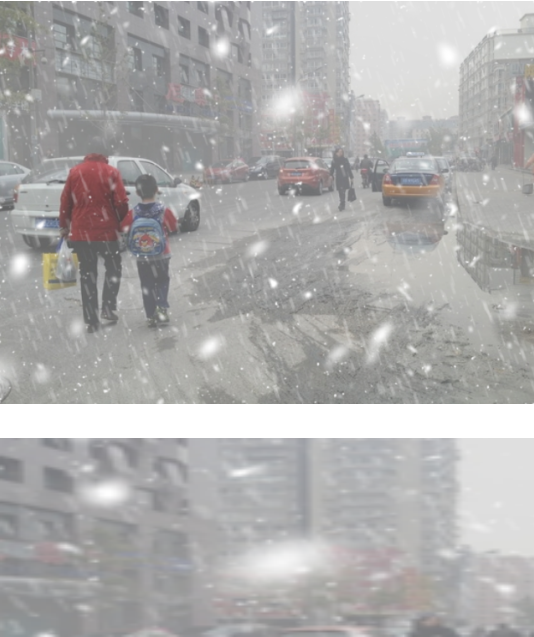} &\hspace{-4mm}
\includegraphics[width = 0.1\linewidth]{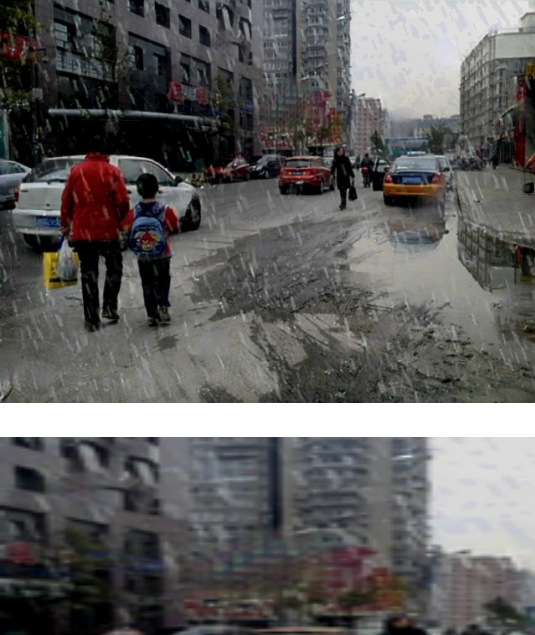} &\hspace{-4mm}
\includegraphics[width = 0.1\linewidth]{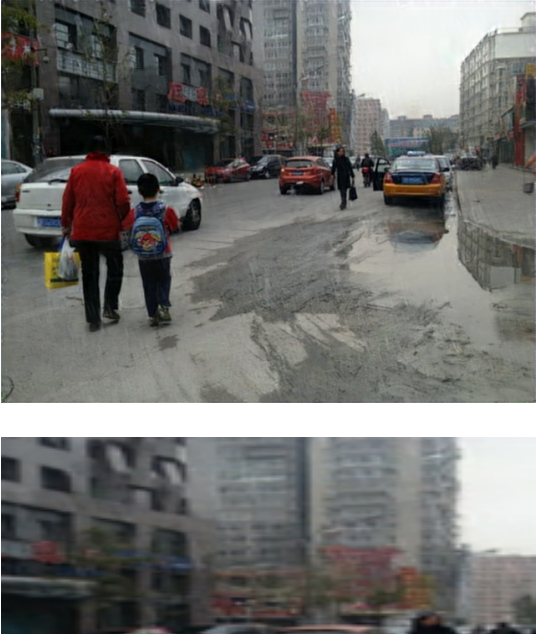} &\hspace{-4mm}
\includegraphics[width = 0.1\linewidth]{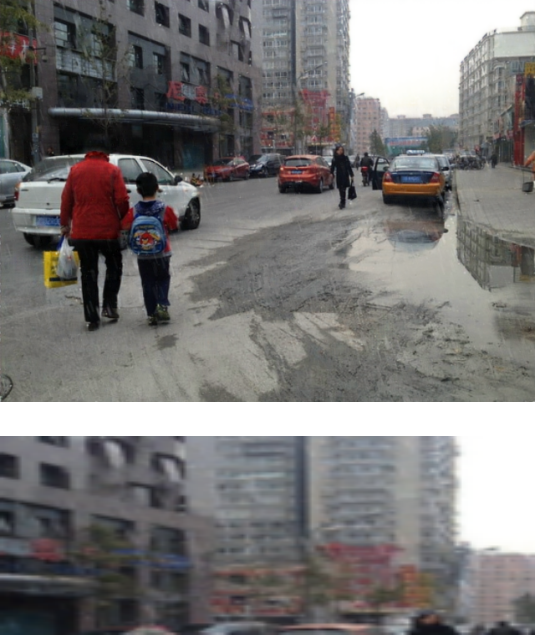} &\hspace{-4mm}
\includegraphics[width = 0.1\linewidth]{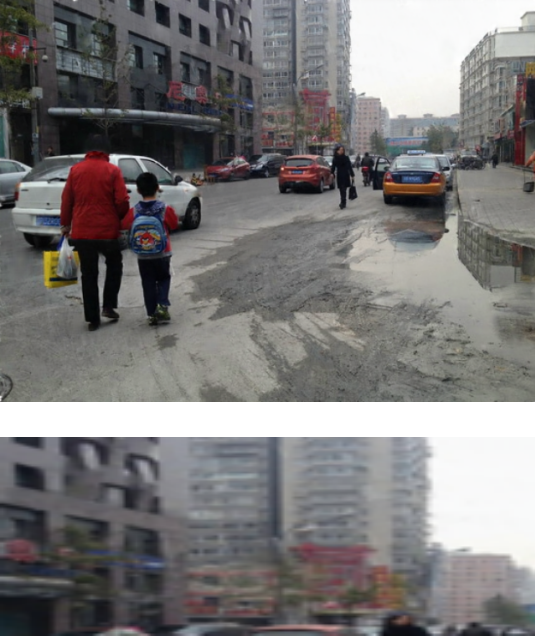} &\hspace{-4mm}
\includegraphics[width = 0.1\linewidth]{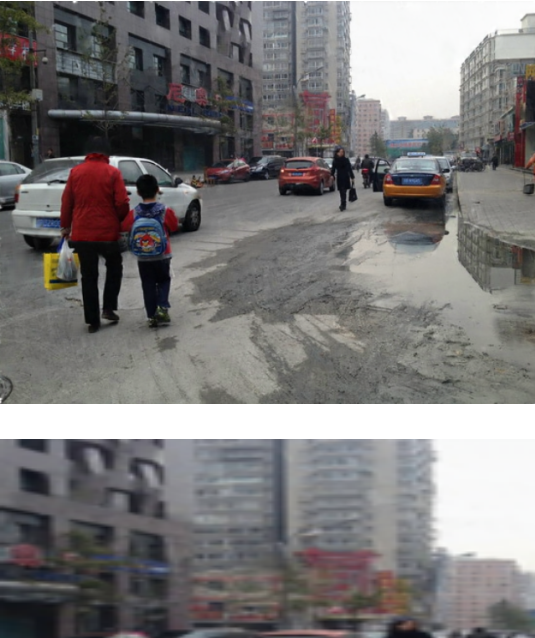} &\hspace{-4mm}
\\
\scriptsize{(a)Input}  &\hspace{-1.5mm} 
\scriptsize{(b)JSTASR} &\hspace{-1.5mm} 
\scriptsize{(c)HDCW-Net} &\hspace{-1.5mm}  
\scriptsize{(d)TransWeather} &\hspace{-1.5mm}
\scriptsize{(e)Ours} &\hspace{-1.5mm}
\scriptsize{(f)Ground-truth} &\hspace{-1.5mm}\\
        \end{tabular}
    \end{center}
\caption{\small Visual comparison of our method (MSP-Former) and the SOTA methods on the synthetic dataset CSD~\cite{hdcwnet}.}\label{fig:visualcomparison1}
\end{figure*}

\noindent\textbf{Parameters and Computational Complexity Analysis}. For an excellent model, parameters and computation are also within the scope of measurement. The number of parameters determines whether the model is lightweight, and the amount of calculation measures the model's efficiency. We present the comparison results of computation of 256$\times$256 size and parameters with state-of-the-art snow removal method in Table \ref{snowresults}. We notice that our MSP-Former is better than the previous SOTA approaches in terms of both parameters and the amount of calculation, which is only 2.83M and 4.26GMacs. We outperform HDCW-Net~\cite{hdcwnet} and TransWeather~\cite{valanarasu2022transweather} in terms of performance-parameter trade-off.


\begin{figure}[!t]

\centering
\includegraphics[width=0.48\textwidth]{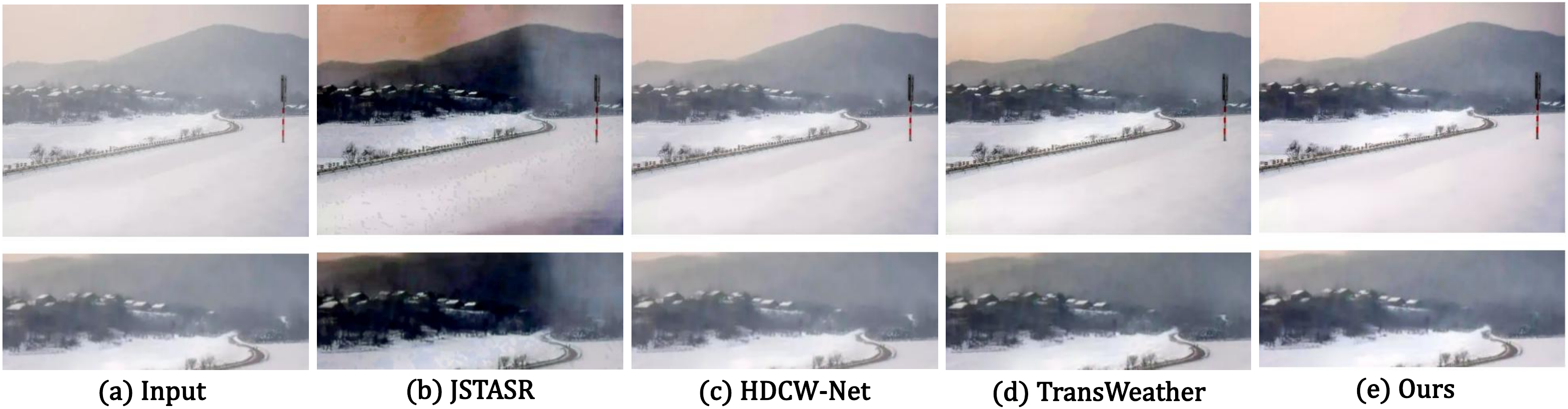} 
\caption{\small Visual comparison of desnowing on the real-world snow image. 
}\label{realcomparison}

\end{figure}
\noindent\textbf{Visual Comparison}. We compare the visual performance of MSP-Former and other state-of-the-art methods for snow removal on desnowing datasets and real snow image, which are presented in Fig.\ref{fig:visualcomparison1} and \ref{realcomparison}. It can be seen that the previous desnowing methods are not able to completely remove all snow scenes at one time due to ignored context information, especially small snow spots and snow marks. Instead, MSP-Former can remove snow degradation of various sizes very well, superior to previous SOTA methods in detail and background restoration. In particular, our method also attracts certain advantages in the recovery of fine snow points and overall color regions.


\vspace{-0.4cm}
\subsection{Ablation Study}
\vspace{-0.2cm}
 For the ablation study, we employ the Charbonnier loss~\cite{charbonnier1994two} as our loss function and train on the CSD~\cite{hdcwnet} dataset with 256$\times$256 for 200 epochs to test the effectiveness of our model. 

\noindent\textbf{Effectiveness of MSP-Self Attention}. To verify the effect of our avgpool local aggregate projection self-attention(AA), we replace our scheme with spatial-reduction attention(SRA) proposed by PVT~\cite{PVT} and max-pooling aggregate projection(MA). The results are shown in Table \ref{MSP ab}.
Experiments demonstrate avgpooling attracts the best performance in targeting various snow degradations.
In addition, We also conduct multi-scale projection(MSP) and single-scale projection(SSP) ablation experiments to verify that our MSP Self-Attention can achieve non-trivial effects on snow removal due to its adaptation to various scales of snow degradation information.
\begin{table}[!h]
\centering
\setlength{\abovecaptionskip}{0cm} 
\setlength{\belowcaptionskip}{-0.5cm}
\caption{\small Ablation Study on the MSP-Self Attention.}\label{MSP ab}
\resizebox{4cm}{!}{
\renewcommand\arraystretch{1.1}
\begin{tabular}{ccccccc}
\toprule[1.2pt]
 Model & \multicolumn{2}{c}{\#Param} & \multicolumn{2}{c}{\#GFlops} & PSNR & SSIM  \\\hline
 SRA & \multicolumn{2}{c}{3.28M} & \multicolumn{2}{c}{4.61G} & 29.84 & 0.931\\
 MA &  \multicolumn{2}{c}{2.83M} & \multicolumn{2}{c}{4.42G} & 28.57 & 0.903 \\
\hline
 SSP & \multicolumn{2}{c}{2.83M} & \multicolumn{2}{c}{4.39G} & 29.49& 0.921\\
 MSP(Ours) & \multicolumn{2}{c}{2.83M} & \multicolumn{2}{c}{4.42G} & \underline{30.12} & \underline{0.935}\\
\bottomrule[1.2pt]
\end{tabular}}
\end{table}

\noindent\textbf{Effectiveness of LCB Module.} We verify the effect of our LCB module in ablation experiments. We designed three schemes to conduct our experiments. As described in Table \ref{LCB ab}, we observe that parallel LCB modules can boost the model's overall recovery performance, which indicates that the model can simultaneously focus on global scene reconstruction and fine-grained recovery of local details. It is worth describing that we found adding channel attention to the convolution module can play a vital role in the overall representative ability of the architecture with only a small amount of parameters and computation.
\begin{table}[!h]
 \setlength{\abovecaptionskip}{-2cm} 
 \setlength{\belowcaptionskip}{-1cm}
\centering
\setlength{\abovecaptionskip}{0.7cm}
\caption{\small Ablation Study on the LCB Module and channel shuffle.}\label{LCB ab}
\resizebox{4cm}{!}{
\renewcommand\arraystretch{1.1}
\begin{tabular}{ccccccc}
\toprule[1.2pt]
 Model & \multicolumn{2}{c}{\#Param} & \multicolumn{2}{c}{\#GFlops} & PSNR & SSIM  \\\hline
 w/o LCB & \multicolumn{2}{c}{2.60M} & \multicolumn{2}{c}{4.26G} & 29.21 & 0.919\\
 LCB w/o CA &  \multicolumn{2}{c}{2.76M} & \multicolumn{2}{c}{4.42G} & 29.61 & 0.923 \\\hline
  w/o CS & \multicolumn{2}{c}{2.83M} & \multicolumn{2}{c}{4.26G} & 29.46 & 0.923\\
 LCB(Ours) & \multicolumn{2}{c}{2.83M} & \multicolumn{2}{c}{4.26G} & \underline{30.12}&\underline{0.935}\\
\bottomrule[1.2pt]
\end{tabular}}
\end{table}

\noindent\textbf{Effectiveness of Channel Shuffle}. We remove and add channel shuffle(CS) operations respectively to observe the changes in model performance, which are described in Table \ref{LCB ab}. The experiment demonstrates that the information reorganization between channels brought by the channel shuffle operations can improve the effects of the network.




\vspace{-0.5cm}
\section{Conclusion}

This paper proposes a single-image snow removal method with a multi-scale projection transformer. Specifically, we employ multi-scale projections to target diverse degradations and combine global context information to promote the reconstruction of the clean scene via multi-head self-attention. On the other hand, we exploit our designed lightweight local capture module for local feature mining. The clever parallel combination of the two can achieve state-of-the-art performance on the snow removal task with a meager amount of parameters and computation, achieving a performance-parameters trade-off.


\vfill\pagebreak

\bibliographystyle{IEEEbib}
\footnotesize{\bibliography{strings,refs}}

\end{document}